\documentclass[runningheads,a4paper]{llncs}

\usepackage{graphicx} 
\usepackage{subfigure} 

\usepackage{amsmath}
\usepackage{amssymb}

\usepackage{graphicx}

\usepackage{url}
\urldef{\mailsa}\path|{alfred.hofmann, ursula.barth, ingrid.haas, frank.holzwarth,|
\urldef{\mailsb}\path|anna.kramer, leonie.kunz, christine.reiss, nicole.sator,|
\urldef{\mailsc}\path|erika.siebert-cole, peter.strasser, lncs}@springer.com|

\newcommand{\bmx}[0]{\begin{bmatrix}}
\newcommand{\emx}[0]{\end{bmatrix}}

\newcommand{\vect}[1]{\mathbf{#1}}

\newcommand{\matr}[1]{\mathbf{#1}}

\newcommand{\vc}[0]{\vect{c}}
\newcommand{\vh}[0]{\vect{h}}

\newcommand{\vx}[0]{\vect{x}}

\newcommand{\vm}[0]{\vect{m}}

\newcommand{\mW}[0]{\matr{W}}

\newcommand{\mV}[0]{\matr{V}}

\newcommand{\LL}[0]{\mathcal{L}}

\newcommand{\RR}[0]{\mathbb{R}}

\newcommand{\E}[0]{\mathbb{E}}

\DeclareMathOperator*{\argmax}{\arg \max}

\begin{document}

\mainmatter  

\title{On the Equivalence Between Deep NADE and Generative Stochastic Networks}


%
%
\author{Li Yao \and 
Sherjil Ozair \and
Kyunghyun Cho \and
Yoshua Bengio$^\star$}
%

\institute{D\'{e}partement d'Informatique et de Recherche Op\'{e}rationelle\\
Universit\'{e} de Montr\'{e}al\\
($\star$) CIFAR Fellow}

%
%

\maketitle

\begin{abstract}
Neural Autoregressive Distribution Estimators (NADEs) have recently been
shown as successful alternatives for modeling high dimensional multimodal
distributions. One issue associated with NADEs is that they rely on a
particular order of factorization for $P(\vx)$.  This issue has been
recently addressed by a variant of NADE called Orderless NADEs and its
deeper version, Deep Orderless NADE.  Orderless NADEs are trained based on
a criterion that stochastically maximizes $P(\vx)$ with all possible orders
of factorizations. Unfortunately, ancestral sampling from deep NADE is very
expensive, corresponding to running through a neural net separately
predicting each of the visible variables given some others.  This work
makes a connection between this criterion and the training criterion for
Generative Stochastic Networks (GSNs).  It shows that training NADEs in
this way also trains a GSN, which defines a Markov chain associated with
the NADE model.  Based on this connection, we show an alternative way to
sample from a trained Orderless NADE that allows to trade-off computing
time and quality of the samples: a 3 to 10-fold speedup (taking into
account the waste due to correlations between consecutive samples of the
chain) can be obtained without noticeably reducing the quality of the
samples. This is achieved using a novel sampling procedure for GSNs called
annealed GSN sampling, similar to tempering methods that combines fast
mixing (obtained thanks to steps at high noise levels) with accurate
samples (obtained thanks to steps at low noise levels).
\end{abstract}

\section{Introduction}

Unsupervised representation learning and deep learning have
progressed rapidly in recent years
\cite{Bengio-Courville-Vincent-TPAMI2013}.  On one hand,
supervised deep learning algorithms have achieved great success.
The authors of \cite{Krizhevsky-2012}, for instance, claimed the
state-of-the-art recognition performance in a challenging object
recognition task using a deep convolutional neural network.  Despite
the promise given by supervised deep learning, its unsupervised
counterpart is still facing several challenges
 \cite{Bengio-chapterSLSP-2013-small}. A large
proportion of popular unsupervised deep learning models are based
on either directed or undirected graphical models with latent
variables \cite{Hinton-Science2006,Hinton06,Salakhutdinov2009}.
One problem of these unsupervised models is that 
it is often intractable to compute the likelihood of a model
exactly.

The Neural Autoregressive Distribution Estimator (NADE) was proposed
in \cite{Larochelle+Murray-2011} to avoid this problem of
computational intractability. 
It was inspired by the early work in \cite{Bengio+Bengio-NIPS99},
which like NADE modeled a binary distribution by decomposing it into a
product of multiple conditional distributions of which each is
implemented by a neural network, with
parameters, representations and computations shared across all these networks.
These kinds of models therefore implement a fully connected
directed graphical model, in which ancestral sampling of the joint
distribution is simple (but not necessarily efficient when the
number of variables, e.g., pixel images, is large). Consequently,
unlike many other latent variable models, it is possible with such
directed graphical models to compute the exact probability of an observation
tractably.
NADEs have
since been extended to model distributions of continuous
variables in \cite{Benigno-et-al-NIPS2013-small}, called a
real-valued NADE (RNADE) which replaces a Bernoulli distribution
with a mixture of Gaussian distributions for each conditional
probability (see ,e.g., \cite{bishop1994mixture}).
The authors of \cite{Benigno-et-al-arxiv-2013}
proposes yet another variant of NADE, called a Deep NADE, that
uses a {\em deep} neural network to compute the conditional
probability of each variable.  
In order to make learning tractable, they proposed a modified
training procedure that effectively trains an ensemble of
multiple NADEs.

Another thread of unsupervised deep learning is based on the family of
autoencoders (see, e.g., \cite{Vincent-JMLR-2010-small}).  The autoencoder
has recently
begun to be understood as a density estimator
\cite{Alain+Bengio-ICLR2013,Bengio-et-al-NIPS2013-small}. These
works suggest that an autoencoder trained with some arbitrary
noise in the input is able to learn the distribution of either
continuous or discrete random variables. This perspective on
autoencoders has been further extended to a generative stochastic
network (GSN) proposed in \cite{Bengio+Laufer-arxiv-2013}. 

Unlike a more conventional approach of directly estimating the
probability distribution of data, a GSN aims to learn a
transition probability of a Markov Chain Monte Carlo (MCMC)
sampler whose stationary distribution estimates the data
generating distribution. The authors of \cite{Bengio+Laufer-arxiv-2013} were able to show
that it is possible to learn the distribution of data with a GSN
having a network structure inspired by a deep Boltzmann machine
(DBM)
\cite{SalHinton07} using this approach. Furthermore, a recently
proposed multi-prediction DBM (MP-DBM)
\cite{Goodfellow2013a}, which 
models the joint distribution of data instance and its label, can
be considered a special case of a GSN and achieves state-of-the-art
classification performance on several datasets.

In this paper, we find a close relationship between the deep NADE
and the GSN. We show that training a deep NADE with the
order-agnostic (OA) training procedure
\cite{Benigno-et-al-arxiv-2013} can be cast as GSN training.
This equivalence allows us to have an alternative theoretical
explanation of the OA training procedure. Also, this allows an
alternative sampling procedure for a deep NADE based on a MCMC
method, rather than ancestral sampling.


In Sec.~\ref{sec:nade} and Sec.~\ref{sec:gsn}, we describe both
NADE and GSN in detail. Based on these descriptions we establish
the connection between the order-agnostic training procedure for
NADE and the training criterion of GSN in
Sec.~\ref{sec:equivalence} and propose a novel sampling algorithm
for deep NADE. In Sec.~\ref{sec:annealed}, we
introduce a novel sampling strategy for GSN called annealed GSN
sampling, which is inspired by tempering methods and does a good
trade-off between computing time and accuracy. We empirically investigate the effect of
the proposed GSN sampling procedure for deep NADE models in
Sec.~\ref{sec:experiments}.

\section{Deep NADE and Order-Agnostic Training}
\label{sec:deep_nade}

In this section we describe the deep NADE and its training
criterion, closely following \cite{Benigno-et-al-arxiv-2013}.

\subsection{NADE}
\label{sec:nade}

NADE \cite{Larochelle+Murray-2011} models a joint distribution
$p(\vx)$ where $\vx \in \RR^D$.  $D$ is the 
dimensionality of $\vx$. NADE factorizes $p(\vx)$ into
\begin{align}
    \label{eq:factorization}
    p(\vx) = \prod_{d=1}^D p(x_{o_d} | \vx_{o_{<d}})
\end{align}
where $o$ is a predefined ordering of $D$ indices.
$o_{<d}$ denotes the first $d-1$ indices of the ordering $o$.

The NADE then models each factor in Eq.~\eqref{eq:factorization}
with a neural network having a single hidden layer $H$. That is,
\begin{align*}
    p(x_{o_d}=1 | \vx_{o_{<d}}) &= \sigma
    (\mV_{.,o_d}\vh_d+b_{o_d}),
\end{align*}
where 
\[
    \vh_d = \phi (\mW_{.,o_{<d}} + \vc).
\]
$\mV \in \RR^{H \times D}$, $b \in \RR^D$, $\mW \in \RR^{H \times
D}$ and $\vc \in \mathcal{R}^H$ are the output weights, the
output biases, the input weights and the hidden biases,
respectively. $\sigma$ is a logistic sigmoid
function, and $\phi$ can be any nonlinear activation function.

To train such a model, one maximizes the log-likelihood function
of the training set
\begin{align}
    \label{eq:ll_nade}
    \theta^* = \argmax_{\theta} \LL_o(\theta) 
    = \argmax_{\theta}
    \frac{1}{N} \sum_{n=1}^N \sum_{d=1}^{D} \log p(x^{n}_{o_d}|
    \vx^{n}_{o_{<d}}, o),
\end{align}
where $\theta$ denotes all the parameters of the model.


\subsection{Deep NADE}
\label{sec:deepnade}

One issue with the original formulation of the NADE is that the
ordering of variables needs to be predefined and fixed.
Potentially, this limits the inference capability of a trained
model such that when the model is asked to infer the conditional
probability which is not one of the factors in the predefined
factorization (See Eq.~\eqref{eq:factorization}).  For instance,
a NADE trained with $D$ visible variables with an
ordering $(1,2,\ldots, D)$, one cannot easily
infer $x_2 \| x_1, x_D$ except by expensive (and intractable)
marginalization over all the other variables.

Another issue is that it is not possible to build a deeper
architecture for NADE with the original formulation without losing
a lot in efficiency. When
there is only a single hidden layer with $H$ units in the neural network
modeling each conditional probability of a NADE, it is possible
to share the parameters (the input weights and the hidden biases)
to keep the computational complexity linear with respect to the
number of parameters, i.e., $O(D H)$. However, if there are more than one
hidden layers, it is not possible to re-use computations in the same
way. In this case, the computational complexity is $O(DH+DH^2L)$ 
where $L$ is the number of hidden layers. Notice the extra $D$
factor in front, compared to the number of parameters which is
$O(DH + H^2 L)$. This comes about because we cannot re-use
the computations performed after the first hidden layer
for predicting the $i$-th variable, when predicting the following
ones. In the one-layer case, this sharing is possible because
the hidden units weighted sums needed when predicting the $i+1$-th variable
are the same as the weighted sums needed when predicting the $i$-th
variable, plus the scalar contributions $w_{ki}$ associated with 
the $k$-th hidden unit and the extra input $x_i$ that is now
available when predicting $x_{i+1}$ but was not available
when predicting $x_i$.


To resolve those two issues, the authors of \cite{Benigno-et-al-arxiv-2013}
proposed the order-agnostic (OA) training procedure that trains a
factorial number of NADEs with shared parameters. In this case,
the following objective function is maximized, instead of $\LL_o$
in Eq.~\eqref{eq:ll_nade}:
\begin{align}
    \label{eq:cost_aonade}
    \LL(\theta) &= \E_{\mathbf{x}^{n}} \sum_{d=1}^{D}
\E_{o_{<d}} \E_{o_{d}} 
\log p(x^{n}_{o_d}| \mathbf{x}^{n}_{o_{<d}}, \theta, o).
\end{align}

This objective function is, however, intractable, since it
involves the factorial number of summations. Instead, in
practice, when training, we use a stochastic approximation
$\widehat{\LL}$ by sampling an ordering $o$, the index of
predicted variable $d$ and a training sample $\vx^{n}$ at each
time:
\begin{align}
    \label{eq:stocost}
    \widehat{\LL}(\theta) =  \frac{D}{D-d+1} \sum_{i \notin o_{<d}} 
    \log p(x^{n}_{i}| \vx^{n}_{o_{<d}}, \theta, o).
\end{align}

Computing $\widehat{\LL}$ is identical to a forward computation
in a regular feedforward neural network except for two
differences. Firstly, according to the sampled ordering $o$, the
input variables of indices $o_{>d}$ are set to $0$, and the
identity of the zeroed indices is provided as extra inputs
(through a binary vector of length $D$). Secondly, the
conditional probabilities of only those variables of indices
$o_{>d}$ are used to compute the objective function
$\widehat{\LL}$. 

This order-agnostic procedure solves the previously raised issues
of the original NADE. Since the model is optimized for all
possible orderings, it does not suffer from being inefficient at
inferring any conditional probability. Furthermore, the lack of
predefined ordering makes it possible to use a single set of
parameters for modeling all conditional distributions. Thus, the
computational cost of training a deep NADE with the OA procedure
is linear with respect to the number of parameters, regardless of
the depth of each neural network.

From here on, we call a NADE trained with the OA procedure simply
a deep NADE to distinguish it from a NADE trained with a usual
training algorithm other than the OA procedure.


\section{Generative Stochastic Networks}
\label{sec:gsn}

In \cite{Bengio-et-al-NIPS2013-small,Bengio+Laufer-arxiv-2013}
a new family of models called generative stochastic
networks (GSN) was proposed, which tackles the problem of modeling a data
distribution, $p(\vx)$, although without providing a tractable
expression for it.

The underlying idea is to learn a transition operator of a
Markov Chain Monte Carlo (MCMC) sampler that samples from the
distribution $p(\vx)$, instead of learning the whole distribution
directly. If we let $p(\vx' \mid \vx)$ be the transition operator, then
we may rewrite it by introducing a latent variable $h$ into
\begin{align} 
    \label{eq:trans}
    p(\vx' \mid \vx) = \sum_{\forall \vh} p(\vx' \mid \vh) p(\vh
    \mid \vx).
\end{align}
In other words, two separate conditional distributions $p(\vx' \mid
\vh)$ and $p(\vh \mid \vx)$ jointly define the transition operator.
In \cite{Bengio-et-al-NIPS2013-small,Bengio+Laufer-arxiv-2013}
it is argued that it is easier to learn these simpler conditional
distributions because they have less modes (they only consider
small changes from the previous state), meaning that the
associated normalization constants can be estimated more
easily (either by an approximate parametrization, e.g., a single
or few component mixture, or by MCMC on a more powerful
parametrization, which will have less variance if the number
of modes is small).


A special form of GSN also found with denoising auto-encoders
predefines $p(\vh \mid \vx)$ such that it does not require learning from data. Then, we
only learn $p( \vx' \mid \vh)$. This is the case in
\cite{Bengio-et-al-NIPS2013-small}, where they proposed to use a
user-defined corruption process, such as randomly masking out
some variables with a fixed probability, for $p(\vh \mid \vx)$. They,
then, estimated $p(\vx' \mid \vh)$ as a denoising autoencoder
$f_{\theta}$, parameterized with $\theta$, that
reverses the corruption process $p(\vh \mid \vx)$
\cite{VincentPLarochelleH2008-small}. 


It was shown in \cite{Bengio-et-al-NIPS2013-small} that if the denoising
process $f_{\theta}$ is a consistent estimator of $p(\vx' \mid \vh)$, this
leads to consistency of the Markov chain's stationary distribution as an
estimator of the data generating distribution.  This is under some
conditions ensuring the irreducibility, ergodicity and aperiodicity of the
Markov chain, i.e., that it mixes.  In other words, training $f_{\theta}$
to match $p(\vx' \mid \vh)$ is enough to learn implicitly the whole
distribution $p(\vx)$, albeit indirectly, i.e., through the definition of a
Markov chain transition operator.  The result from
\cite{Bengio+Laufer-arxiv-2013} further suggests that it is possible to
also parameterize the corruption process $p(\vh \mid \vx)$ and learn both
$p(\vh \mid \vx)$ and $p(\vx' \mid \vh)$ together.


From the qualitative observation on some of the learned
transition operators of GSNs (see, e.g., \cite{Bengio+Laufer-arxiv-2013}), it is clear that the learned
transition operator quickly finds a plausible mode in the whole
distribution, even when the Markov chain was started from a
random configuration of $\vx$. This is because the GSN reconstruction
criterion encourages the learner to quickly move from low probability
configurations to high-probability ones, i.e., to burn-in quickly.
This is in contrast to using a Gibbs
sampler to generate samples from other generative models that
explicitly model the whole distribution $p(\vx)$, which requires
often many more \textit{burn-in} steps before the Markov chain finds a
plausible mode of the distribution.

\section{Equivalence between deep NADE and GSN}
\label{sec:equivalence}

Having described both deep NADE and GSN, we now establish the
relationship, or even equivalence, between them. In particular,
we show in this section that the order-agnostic (OA) training
procedure for NADE is one special case of GSN learning.

We start from the stochastic approximation to the objective
function of the OA training procedure for deep NADE in
Eq.~\eqref{eq:stocost}. We notice that the sampled ordering $o$
in the objective function $\hat{\LL}$ can be replaced with
another random variable $\vm \in \left\{ 0, 1\right\}^D$, where $D$
is the dimensionality of an observation $x$. The binary mask
$\vm$ is constructed such that 
\[
    m_i = \left\{ 
        \begin{array}{l l}
            1,&\text{if } i \in o_{<d} \\
            0,&\text{otherwise }
        \end{array}
        \right.
\]

Then, we rewrite Eq.~\eqref{eq:stocost} by
\begin{align}
    \label{eq:deriv1}
    \widehat{\LL}(\theta) \propto& \sum_{i=1}^D
    (1 - m_i) \log p(x^{n}_i \mid \vm \odot \vx^n, \theta, \vm) 
    \nonumber \\
    =& \sum_{i=1}^D
    \log \left( 
        m_i + (1 - m_i) p(x^{n}_i \mid \vm \odot \vx^n, \theta, \vm)
    \right) 
    \nonumber \\
    =& \sum_{i=1}^D
    \log \left( 
        m_i + (1 - m_i) p(x^{n}_i \mid \vh^{(n)}, \theta)
    \right)
\end{align}
where $m_i$ is the $i$-th element of the binary mask $\vm$, and
$\odot$ is an element-wise multiplication. We introduced a new
variable $\vh = \left[ \vm, \vm \odot \vx^n \right] \in \RR^{2D}$ which is a
concatenation of the corrupted copy (some variables masked
out) of $\vx^n$ and the sampled mask $\vm$.

It is now easy to see the connection between the objective
function of the OA training procedure in Eq.~\eqref{eq:deriv1} to
a GSN training criterion using a user-defined (not learned)
corruption process which we described in the earlier section. 

In this case, the corruption process $p(\vh \mid \vx)$
(Eq.~\eqref{eq:trans}) is 
\begin{align}
    \label{eq:h_x_gsn_nade}
    p(\vh \mid \vx) = p(\left[\vm, \vm \odot \vx\right] \mid \vx) 
    = \prod_{i=1}^D k \prod_{j=1}^D
    \delta_{m_j x_j}(h_{j+D}),
\end{align}
where $k$ is a random number sampled uniformly between 0 and 1, and
$\delta_{\mu}(a)$ is a shifted Dirac delta function which is $1$ only when
$a=\mu$ and 0 otherwise. This means that sampling is done by first
generating an uniformly random binary mask $\vm$ and then taking $\vm \odot
\vx$ as the corrupted version of $\vx$.

The conditional probability of $\vx'$ given $\vh$ is
\begin{align}
    \label{eq:x_h_gsn_nade}
    p(\vx'\mid\vh)=\prod_{i=1}^D \left[ m_i\delta_{x_i}(x'_i) + 
    (1-m_i) p(x'_i\mid r_i(\vx \odot \vm\mid \theta))\right],
\end{align}
where $r_i$ is a parametric function (neural network) that models
the conditional probability.

If we view the estimation of $p(\vx'\mid\vh)$ in Eq.~\eqref{eq:x_h_gsn_nade}
as a denoising autoencoder, one effectively ignores each
variable $x'_i$ with its mask $m_i$ set to $1$, since the sample
of $x'_i$ from Eq.~\eqref{eq:x_h_gsn_nade} is always $x_i$
due to $\delta_{x_i}(x'_i)$. A high-capacity auto-encoder could learn
that when $m_i=1$, it can just copy the $i$-th input to the $i$-th
output. On the other hand, when $m_i$ is $0$,
training this denoising autoencoder would maximize $\log p(x_i' \mid
r_i(\vx \odot \vm\mid \theta))$, making it assign high probability to
the original $x_i$ given the non-missing inputs. Therefore, maximizing the logarithm
of $p(\vx'\mid\vh)$ in Eq.~\eqref{eq:x_h_gsn_nade} is equivalent
to maximizing $\widehat{L}$ in Eqs.~\eqref{eq:deriv1} and
\eqref{eq:stocost} up to a constant.

In essence, maximizing $\widehat{\LL}$ in Eq.~\eqref{eq:stocost}
is equivalent to training a GSN with the conditional
distributions defined in
Eqs.~\eqref{eq:h_x_gsn_nade}--\eqref{eq:x_h_gsn_nade}.
Furthermore, the chain defined in this way is ergodic as every
state $\vx$ has a non-zero probability at each step ($\vx \to
\vx'$), making this GSN chain a valid MCMC sampler.

\subsection{Alternative Sampling Method for NADE}
\label{sec:gsn_nade_sample}

Although the training procedure of the deep NADE introduced in
\cite{Benigno-et-al-arxiv-2013} is order-agnostic, sampling from
the deep NADE is not. 

The authors of \cite{Benigno-et-al-arxiv-2013} proposed an ancestral sampling
method for a deep NADE. Firstly, one randomly selects an ordering
uniformly from all possible orderings. One generates a sample of
each variable from its conditional distribution following the
selected ordering. When $D$, $H$ and $L$ are respectively the
dimensionality of the observation variable, the number of hidden
units in each hidden layer and the number of layers, the time
complexity of sampling a single sample using this ancestral
approach is $O(DLH^2)$.

We propose here an alternative sampling strategy based on our
observation of the equivalence between the deep NADE and GSN.
The new strategy is simply to alternating between sampling from
$p(\vh \mid \vx)$ in Eq.~\eqref{eq:h_x_gsn_nade} and $p(\vx' \mid
\vh)$ in Eq.~\eqref{eq:x_h_gsn_nade}, which corresponds to
performing Markov Chain Monte Carlo (MCMC) sampling on $p(\vx)$.
The computational complexity of a single step ($\vx \to \vh \to
\vx'$) in this case is $O(DH + LH^2)$.

Unlike the original ancestral strategy, the proposed approach
does not generate an exact sample in a single step. Instead, one
often needs to run the chain $K$ steps until the exact,
independent sample from the stationary distribution of the chain
is collected, which we call \textit{burn-in}. In other words, the
new approach requires $O(KDH + KLH^2)$ to collect a single
sample in the worst case.\footnote{
    Since it is a usual practice to collect
    every $t$-th samples from the same chain, where $t << K$, we
    often do not need $KN$ steps to collect $N$ samples.
}

If we assume that $H$ is not too larger than $D$ ($H=O(D)$ or
$H=\Theta(D)$), which is an usual practice, the time complexity
of the ancestral approach is $O(D^3)$, and that of the proposed
GSN approach is $O(KD^2)$, where we further assume that $L$ is a
small constant. Effectively, if the MCMC method used in the
latter strategy requires only a small, controllable number $K$ of
steps to generate a single exact, independent sample such that $K
\ll D$, the new approach is more efficient in collecting samples
from a trained deep NADE.  Importantly, as we have already
mentioned earlier, a GSN has been shown to learn a transition
operator of an MCMC method that requires only a small number of
burn-in steps. 

In the experiments, we investigate empirically
whether this new sampling strategy is computationally more
efficient than the original ancestral approach in a realistic
setting.



\subsection{The GSN Chain Averages an Ensemble of Density Estimators}

As discussed in \cite{Benigno-et-al-arxiv-2013}, maximizing
$\widehat{\LL}$ in Eq.~\eqref{eq:cost_aonade} can be considered
as training a factorial number of different NADEs with shared
parameters. Each NADE differs from each other by the choice of
the ordering of variables and may assign a different probability
to the same observation.\footnote{
    The fact that all ensembles share the exact same parameters
    makes it similar to the recently proposed technique of
    dropout \cite{Hinton-et-al-arxiv2012}.
} 
Based on this observation, it is suggested in \cite{Benigno-et-al-arxiv-2013}
to use the average of the assigned probabilities by all
these NADE, or a small randomly chosen subset of them, as the
actual probability.

This interpretation of seeing the deep NADE as an ensemble of
multiple NADEs and our earlier argument showing that the deep
NADE training is special case of GSN training naturally leads to a question:
\textit{does the GSN Markov chain average an ensemble of density/distribution
estimators?}. 

We claim that the answer is yes. 
From the equivalence we showed in this paper, it is clear that a
GSN trained with a criterion such as the NADE criterion
\textit{learns} an ensemble
of density/distribution estimators (in this case, masking noise
with the reconstruction conditional distribution in
Eq.~\eqref{eq:x_h_gsn_nade}). Furthermore, when one samples
from the associated GSN Markov chain, one is averaging the
contributions associated with different orders. So, although
each of these conditionals (predicting a subset given another
subset) may not be consistent with a single joint distribution,
the associated GSN Markov chain which combines them randomly
does define a clear joint distribution: the stationary
distribution of the Markov chain. Clearly, this stationary
distribution is an ensemble average over all the possible
orderings.

\begin{figure*}[t]
\centering
\includegraphics[width=0.99\textwidth]{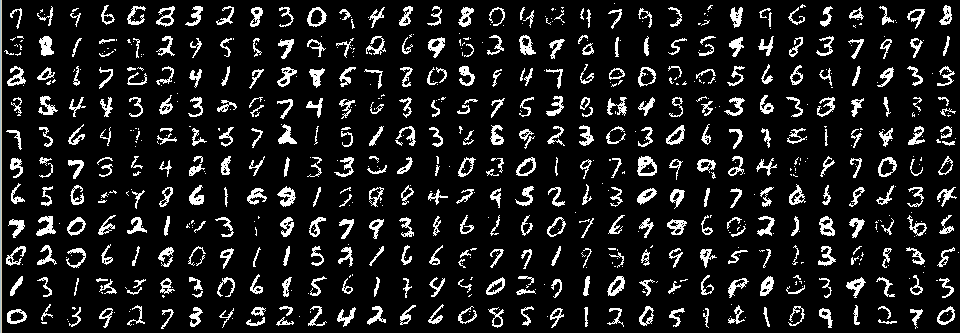}
\caption{
Independent samples generated by the ancestral sampling procedure from the
deep NADE.
}
\label{fig:nade_samples}
\end{figure*}

\section{Annealed GSN Sampling}
\label{sec:annealed}

With the above proposal for GSN-style sampling of a Deep NADE model, one
can view the average fraction $p$ of input variables that are resampled at
each step as a kind of noise level, or the probability of resampling any
particular visible variable $x_i$.  With uniform sampling of subsets, we
obtain $p=0.5$, but both higher and lower values are possible.  When $p=1$,
all variables are resampled independently and the resulting samples are
coming from the marginal distributions of each variable, which would be a
very poor rendering of the Deep NADE distribution, but would mix very
well. With $p$ as small as possible (or more precisely, resampling only one randomly
chosen variable given the
others), we obtain a {\em Gibbs sampler} associated with the Deep NADE
distribution, which we know has the same stationary distribution as Deep
NADE itself. However, this would mix very slowly and would not bring any
computational gain over ancestral sampling in the Deep NADE model (in fact
it would be considerably worse because the correlation between consecutive
samples would reduce the usefulness of the Markov chain samples, compared
to ancestral sampling that provides i.i.d. samples).  With intermediate values
of $p$, we obtain a compromise between the fast computation and
the quality of samples.

However, an even better trade-off can be reached by adopting a form
of annealed sampling for GSNs, a general recipe for improving the
compromise between accuracy of the sampling distribution and mixing
for GSNs. For this purpose we talk about a generic noise level,
although in this paper we refer to $p$, the probability of resampling
any particular visible variable.

The idea is inspired by annealing and tempering methods that have been
useful for undirected graphical models~\cite{Neal94b,Neal-2001}: 
{\em before sampling from the low-noise regime, we run the high-noise
version of the transition operator and gradually reduce the noise
level over a sequence of steps.} The steps taken at high noise allow to mix quickly while
the steps taken at low noise allow to burn-in near high probability
samples. Therefore we consider an approximation of the GSN transition
operator which consists of the successive application of a sequence
of instances of the operator associated with gradually reduced
noise levels, ending at the target noise level. Conceptually, it is
as if the overall Markov chain was composed, for each of its steps,
by a short chain of steps with gradually decreasing noise levels.
By making the annealing schedule have several steps at or near
the target low noise level, and by controlling the lengths of
these annealing runs, we can trade-off between accuracy of the
samples (improved by a longer annealing run length) and speed
of computation.

In the experiments, we used the following annealing schedule:
\[
  p_t = \max(p_{\min}, p_{\max} - (t - 1) * (p_{\max} - p_{\min})
  / (\alpha * (T - 1)))
\]
where $p_{\max}$ is the high noise level, $p_{\min}$ is the low
(target) noise level, $T$ is the length of the annealing run, and
$\alpha\geq 1$ controls which fraction of the run is spent in
annealing vs doing burn-in at the low noise level.

\begin{figure}[tp]
\centering
\begin{minipage}{0.48\textwidth}
\includegraphics[width=0.99\columnwidth]{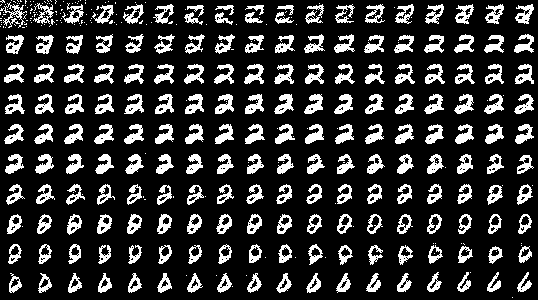}
\end{minipage}
\hfill
\begin{minipage}{0.48\textwidth}
\includegraphics[width=0.99\columnwidth]{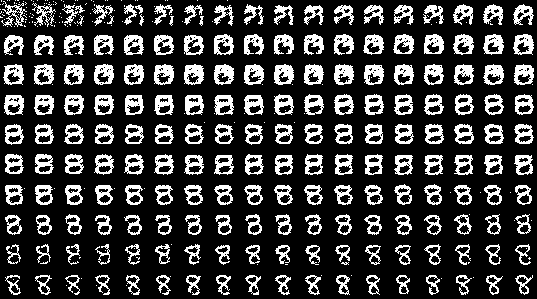}
\end{minipage}
\caption{
The consecutive samples from the two independent GSN sampling
chains without any annealing strategy. Both chains started from
uniformly random configurations. Note the few spurious samples
which can be 
avoided with the annealing strategy (see Figure~\ref{fig:annealed-samples}).
}
\label{fig:multimodal-example}
\end{figure}


\section{Experiments}
\label{sec:experiments}

\subsection{Settings: Dataset and Model}

We run experiments using the handwritten digits dataset (MNIST, 
\cite{LeCun+98}) which has 60,000 training samples and
10,000 test samples. Each sample has 784 dimensions, and we
binarized each variable by thresholding at $0.5$. The training
set is split into two so that the first set of 50,000 samples is
used to train a model and the other set of 10,000 samples is used
for validation.

Using MNIST we trained deep NADE with various architectures and
sets of hyperparameters using the order-agnostic (OA) training
procedure (see Sec.~\ref{sec:deepnade}). The best deep NADE model
according to the validation performance has two hidden layers
with size 2000 and was trained with a linearly decaying learning
rate schedule (from $0.001$ to $0$) for 1000 epochs. We use this
model to evaluate the two sampling strategies described and
proposed earlier in this paper.

\subsection{Qualitative Analysis}

Fig.~\ref{fig:nade_samples} shows a subset of 10,000 samples
collected from the deep NADE using the conventional ancestral
sampling. The average log-probability of the samples is $-70.36$
according to the deep NADE. As each sample by the ancestral
sampling is exact and independent from others, we use these
samples and their log-probability as a baseline for assessing the
proposed GSN sampling procedure.

We first generate samples from the deep NADE using the GSN
sampling procedure without any annealing strategy. A sampling
chain is initialized with a uniformly random configuration, and a
sample is collected at each step. The purpose of this sampling is
to empirically confirm that the GSN sampling does not require
many steps for burn-in. We ran two independent chains and
visualize the initial 240 samples from each of them in
Fig.~\ref{fig:multimodal-example}, which clearly demonstrates
that the chain rapidly finds a plausible mode in only a few
steps.

Although this visualization suggests a faster burn-in, one
weakness is clearly visible from these figures
(Fig.~\ref{fig:multimodal-example}. The chain generates many
consecutive samples of a single digit before it starts generating
samples of another digit.  That is, the samples are highly
correlated temporally, suggesting potentially slow convergence to
the stationary distribution.

\begin{figure*}[t]
\centering
\includegraphics[width=0.9\textwidth]{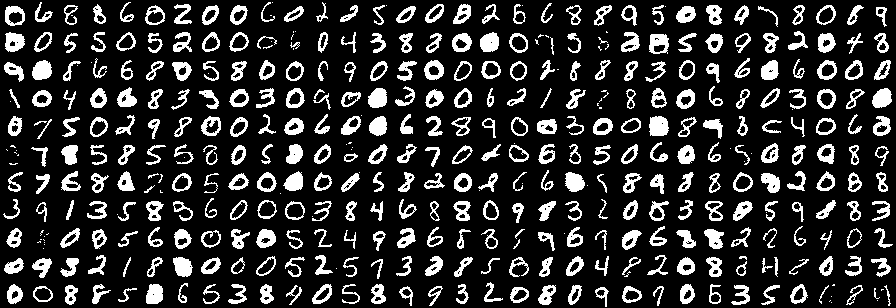}
\caption{
Samples generated by the annealed GSN sampling procedure for the
same deep NADE model. Visually the quality is comparable to the ancestral
samples, and mixing is very fast. This is obtained with $p_{\max}=0.9$,
$p_{\min}=0.1$, $\alpha=0.7$ and $T=20$.
}
\label{fig:annealed-samples}
\end{figure*}

We then tried sampling from the deep NADE using the novel
annealed GSN sampling proposed in Sec.~\ref{sec:annealed}.
Fig.~\ref{fig:annealed-samples} visualizes the collected, samples
over the {\em consecutive} annealing runs. Compared to the
samples generated using the ordinary GSN sampling method, the
chain clearly mixes well. One can hardly notice a case where a
successive sample is a realization of the same digit from the
previous sample.  Furthermore, the samples are qualitatively
comparable to those exact samples collected with the ancestral
sampling (see Fig.~\ref{fig:nade_samples}).

In the following section, we further investigate the proposed
annealed GSN sampling in a more quantitative way, in comparison
to the ancestral sampling.

\subsection{Quantitative Results}

We first evaluate the effect of using a user-defined noise level
in $p(\vh \mid \vx)$ (Eq.~\eqref{eq:h_x_gsn_nade}). We generated
1000 samples from GSN chains with five different noise levels;
$0.1$, $0.3$, $0.4$, $0.5$ and $0.6$. For each noise level, we
ran 100 independent chains and collected every 200-th sample from
each chain. As a comparison, we also generated 1000 samples from
a chain with the proposed annealed GSN sampling with
$p_{\max}=0.9$, $p_{\min}=0.1$ and $\alpha=0.7$.

We computed the log-probability of the set of samples collected
from each chain with the deep NADE to evaluate the quality of the
samples.  Tab.~\ref{tab:non_anneal_different_p} lists the
log-probabilities of the sets of samples, which clearly shows
that as the noise level increases the quality of the samples
degrades. Importantly, none of the chains were able to generate
samples from the model that are close to those generated by the
ancestral sampling. However, the annealed GSN sampling was able
to generate samples that are quantitatively as good as those from
the ancestral sampling.

\begin{table}[ht]
    \begin{minipage}{0.4\textwidth}
\centering
\begin{tabular}{c | c}
\hline
Noise  & Log-Probability \\
\hline
\hline
0.1    &  -77.1  \\
0.3    & -78.93  \\
0.4    & -77.9 \\
0.5    & -81.1 \\
0.6    & -88.1 \\
Annealed &  -69.72 \\
\hline
Ancestral & -70.36 \\
\hline 
\end{tabular}
\end{minipage}
\hfill
\begin{minipage}{0.58\textwidth}
\caption{Log-probability of 1000 samples when anealing is not used.
To collect samples, 100 parallel chains are run and 10 samples are taken from 
each chain and combined together. The noise level is fixed at a particular level 
during the sampling. We also report the best log-probability of samples generated 
with an annealed GSN sampling.
}
\label{tab:non_anneal_different_p}
\end{minipage}
\end{table}

\begin{figure}[ht]
    \begin{minipage}{0.48\textwidth}
\centering
\includegraphics[width=0.99\columnwidth]{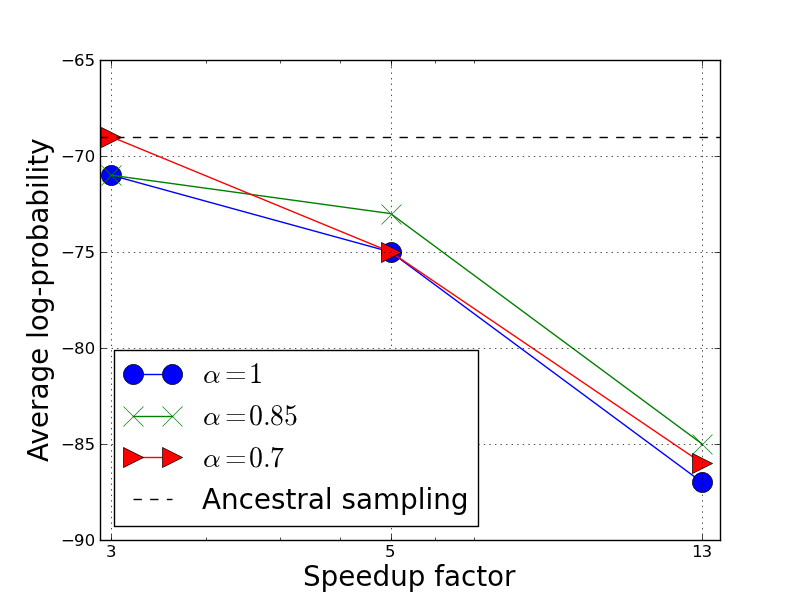}
\end{minipage}
\hfill
\begin{minipage}{0.48\textwidth}
\caption{
The annealed GSN sampling procedure is compared against NADE ancestral sampling, trading off
the computational cost (computational wrt to ancestral sampling on x-axis) against
log-likelihood of the generated samples (y-axis). The computational cost discards
the effect of the Markov chain autocorrelation by estimating the effective number of samples
and increasing the computational cost accordingly.
}
\label{fig:factor}
\end{minipage}
\end{figure}

We also perform quantitative analysis to measure the computational gain when using 
the GSN sampling procedure to generate samples. The speedup by using annealed GSN 
sampling instead of ancestral sampling is shown in Figure \ref{fig:factor}. To compute the speedup factor, 
we timed both the ancestral NADE sampling and GSN sampling on the same machine 
running single process. NADE sampling takes 3.32 seconds per sample and GSN sampling 
takes 0.009 seconds. That means the 
time to get one sample in ancestral sampling can get 369 samples in GSN sampling. Although the the direct speedup factor is 369, it must be discounted
because of the autocorrelation of successive samples in the GSN chain.  
Then we perform different GSN sampling runs with different settings of $\alpha$. 
Figure \ref{fig:factor} shows the results with different $\alpha$. For each 
$\alpha$, a GSN sampling starting at random is run and we collect one out of 
every $K$ samples till 1000 samples are collected. The effective sample size \cite{geyer1992practical}
is then estimated based on the sum of the autocorrelations in the autocorrelation
factor. The speedup factor is discounted accordingly.

\section{Conclusions}
\label{sec:conclusion}

This paper introduced a new view of the orderless NADE training procedure
as a GSN training procedure, which yields several interesting conclusions:
\begin{itemize}
\item The orderless NADE training procedure also trains a GSN model,
where the transition operator randomly selects a subset of input variables
to be resampled given the others.
\item Whereas orderless NADE models really represent an ensemble of
conditionals that are not all compatible, the GSN interpretation provides
a coherent interpretation of the estimated distribution
through the stationary distribution of the
associated Markov chain.
\item Whereas ancestral sampling in NADE is exact, it is very expensive
for deep NADE models, multiplying computing cost (of running once through
the neural network to make a prediction) by the number of
visible variables. On the other hand, each step of the associated GSN
Markov chain only costs running once through the predictor, but
because each prediction is made in parallel for all the resampled variables,
each such step is also less accurate, unless very few variables are
resampled. This introduces a trade-off between accuracy and computation
time that can be controlled. This was validated experimentally.
\item A novel sampling procedure for GSNs was introduced, called
annealed GSN sampling, which permits a better trade-off by combining
high-noise steps with a sequence of gradually lower noise steps,
as shown experimentally.
\end{itemize}




\section*{Acknowledgments}
We would like to thank the developers of
Theano~\cite{bergstra+al:2010-scipy,Bastien-Theano-2012}.  We would also
like to thank CIFAR, and Canada Research Chairs for funding, and Compute
Canada, and Calcul Qu\'ebec for providing computational resources. 

\bibliography{strings,strings-shorter,ml,aigaion,myref}
\bibliographystyle{abbrv}

\end{document}